\documentclass[11pt]{article}

\usepackage[preprint]{acl}

\usepackage{times}
\usepackage{latexsym}
\usepackage[T1]{fontenc}
\usepackage[utf8]{inputenc}
\usepackage{microtype}
\usepackage{inconsolata}
\usepackage{graphicx}
\usepackage{amsmath}
\usepackage{amssymb}
\usepackage{placeins}
\usepackage{booktabs}
\usepackage{pdflscape}
\usepackage{multirow}
\usepackage{rotating}
\usepackage{afterpage}
\usepackage{multicol}

\usepackage{pifont}
\newcommand{\cmark}{\ding{51}}
\newcommand{\xmark}{\ding{55}}

\usepackage[title]{appendix}

\title{AdaGATE: Adaptive Gap-Aware Token-Efficient Evidence Assembly for Multi-Hop Retrieval-Augmented Generation}

\author{
  Yilin Guo \\
  Center for Data Science \\
  New York University \\
  \texttt{yg3030@nyu.edu}
  \And
  Yinshan Wang \\
  Tandon School of Engineering \\
  New York University \\
  \texttt{yw9023@nyu.edu}
  \And
  Yixuan Wang \\
  Center for Data Science \\
  New York University \\
  \texttt{yw8735@nyu.edu}
}

\begin{document}
\maketitle
\begin{abstract}
Retrieval-augmented generation (RAG) remains brittle on multi-hop questions in realistic deployment settings, where retrieved evidence may be noisy or redundant and only limited context can be passed to the generator. Existing controllers address parts of this problem, but typically either expand context additively, select from a fixed top-\(k\) set, or optimize relevance without explicitly repairing missing bridge facts. We propose \textsc{AdaGATE}, a \textbf{training-free} evidence controller for multi-hop RAG that frames evidence selection as a \textbf{token-constrained repair} problem. AdaGATE combines entity centric gap tracking, targeted micro-query generation, and a utility based selection mechanism that balances \textbf{gap coverage}, \textbf{corroboration}, \textbf{novelty}, \textbf{redundancy}, and \textbf{direct question relevance}. We evaluate AdaGATE on HotpotQA under clean, redundancy, and noise injected retrieval conditions. Across all three settings, AdaGATE achieves the \textbf{best evidence F1} among the compared controllers, reaching 62.3\% on clean data and 71.2\% under redundancy injection, while using \textbf{2.6$\times$ fewer input tokens} than Adaptive-\textit{k}. These results suggest that explicit gap-aware repair, combined with token-efficient evidence selection, improves robustness in multi-hop RAG under imperfect retrieval. Our code and evaluation pipeline are available at \url{https://github.com/eliguo/AdaGATE}.
\end{abstract}

\section{Introduction}

Retrieval-augmented generation (RAG) improves large language models (LLMs) by conditioning generation on external documents, reducing hallucinations and improving factual accuracy \citep{fan2024surveyRAGLLM}. In realistic deployments, however, retrieved evidence is often noisy, redundant, or incomplete. Under limited context budgets imposed by API costs and latency constraints \citep{taguchi2025adaptivek,peng2025adagres}, RAG systems cannot simply pass all retrieved content to the generator. This challenge is especially acute for multi-hop questions, where answering often depends on assembling a small set of complementary passages: missing a bridge fact can cause failure, while admitting redundant or misleading evidence can distort the final answer. These constraints motivate viewing multi-hop RAG as a token-constrained evidence assembly problem under imperfect retrieval.

Prior work has shown that RAG performance depends not only on retrieval quality, but also on how retrieved evidence is selected and organized for generation. Irrelevant passages can substantially degrade answer quality \citep{cuconasu2024powerofnoise}, and dense retrievers often return clusters of nearly duplicate chunks that reduce coverage of the full reasoning chain. Recent methods address different parts of this problem. Self-RAG trains an LLM to interleave generation with retrieval-aware reflection \citep{asai2023selfrag}. Adaptive-$k$ selects a query specific number of passages based on similarity score gaps \citep{taguchi2025adaptivek}. SEAL-RAG performs entity centric gap repair through targeted micro-queries and replacement-based updates \citep{lahmy2025sealrag}. However, these approaches typically either expand context additively, operate over a fixed top-$k$ set, or do not explicitly balance gap repair, redundancy, and context efficiency within a single evidence selection procedure.

\begin{figure*}[!t]
    \centering
    \includegraphics[width=0.95\textwidth]{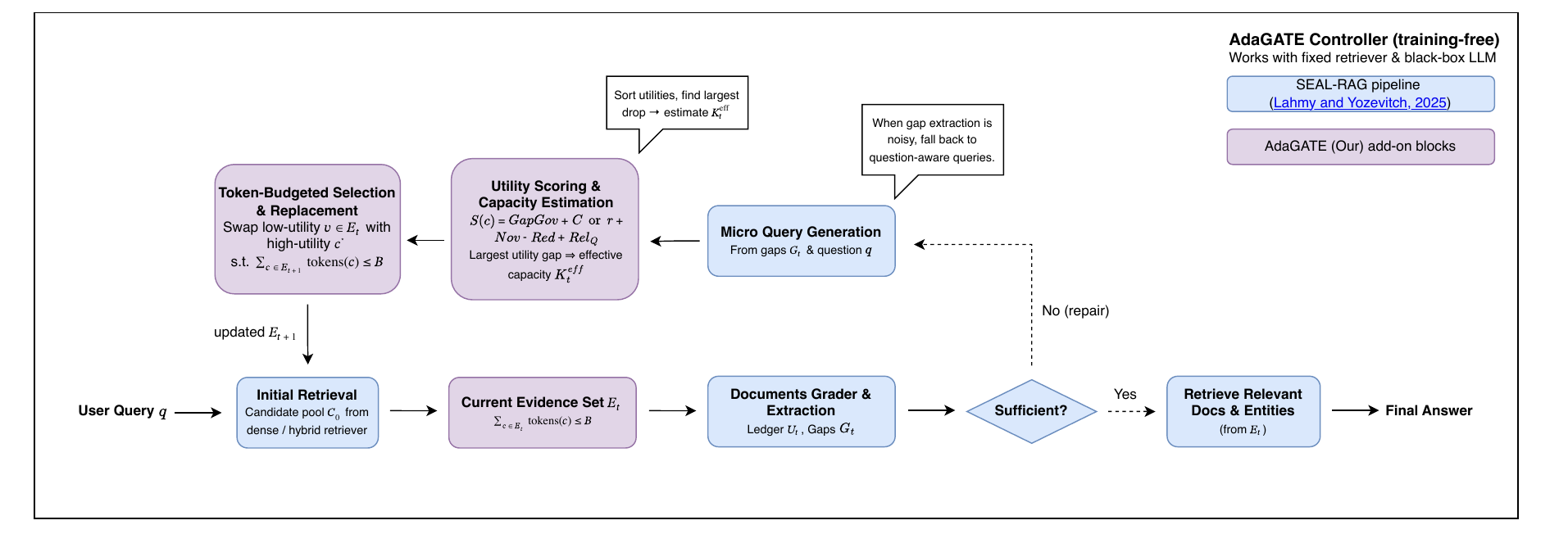}
    \caption{AdaGATE framework overview. Unlike SEAL-RAG \citep{lahmy2025sealrag}, it introduces a training-free, gap-aware controller that explicitly enforces token efficiency on top of a fixed retriever and LLM.}
    \label{fig:AdaGATE_flow}
\end{figure*}

We propose \textsc{AdaGATE} (Figure~\ref{fig:AdaGATE_flow}), a \textbf{training-free controller} that treats multi-hop evidence selection as a \textbf{gap-aware}, \textbf{token-efficient repair}. AdaGATE maintains an entity centric ledger, issues targeted micro-queries with a question aware fallback channel, and \textbf{scores candidates} with a utility function balancing gap coverage, corroboration, novelty, redundancy, and question relevance. A utility adaptive capacity heuristic then assembles a compact evidence set under a global token budget.

Evaluated on HotpotQA under clean, redundancy injection and noise injection conditions against four baselines, AdaGATE achieves the highest evidence F1 across all three settings (62.3\% clean, 71.2\% redundancy, 62.7\% noise) while using 2.6$\times$ fewer input tokens than Adaptive-$k$.

The main contributions of this work are:
(1) We formulate multi-hop RAG under imperfect retrieval as a token constrained evidence repair problem and clarify the limitations of fixed-$k$ evidence controllers in noisy and redundant settings;
(2) We introduce \textsc{AdaGATE}, a training-free controller that combines entity centric gap tracking, utility based evidence scoring, and adaptive capacity control for compact evidence assembly; and
(3) We develop a stress tested evaluation protocol on HotpotQA with controlled redundancy and noise injection, and compare controllers across answer quality, grounding, and token efficiency.

\section{Related Work}

\subsection{Multi-Hop RAG under Imperfect Retrieval}

Standard RAG pipelines retrieve a fixed top-$k$ set of passages and concatenate them with the query, implicitly treating evidence selection as a one-shot step. Multi-hop QA benchmarks such as HotpotQA and 2WikiMultiHopQA expose the limits of this assumption: answering often requires combining complementary facts across documents, and omitting a single bridge passage can cause failure even when retrieval recall is high \citep{yang2018hotpotqa,welbl2018twikimultihopqa}. Prior work further shows that long or noisy contexts degrade generation quality due to distractor sensitivity and ``lost in the middle'' effects \citep{liu2023lostinthemiddle,cuconasu2024powerofnoise}. These findings motivate controllers that explicitly manage evidence composition under imperfect retrieval.

\subsection{Active and Corrective RAG Controllers}

Recent work has made the RAG controller more adaptive. Self-RAG trains an LLM to interleave generation with retrieval awareness reflection \citep{asai2023selfrag}. Adaptive-RAG routes questions among non retrieval, single and multi step strategies based on estimated complexity \citep{jeong2024adaptiverag}. CRAG evaluates retrieved documents and triggers corrective actions such as additional retrieval or document decomposition when evidence quality appears low \citep{yan2024crag}. These methods make retrieval more adaptive, but they primarily decide when or whether to retrieve, and several rely on model finetuning, rather than focusing on token efficient evidence assembly for multi-hop reasoning.

SEAL-RAG is most closely related to our setting \citep{lahmy2025sealrag}. It maintains an entity centric ledger, identifies missing information as gaps, and issues targeted micro-queries to repair a fixed evidence set through replacement rather than expansion. Our work builds directly on this line of explicit gap-aware repair, but extends it beyond a fixed top-$k$ setting.

\subsection{Adaptive Evidence Selection and Position of This Work}

A complementary line of work studies how much context to include. Adaptive-$k$ selects a query specific number of passages by identifying the largest drop in sorted similarity scores \citep{taguchi2025adaptivek}. AdaGReS formulates evidence selection as a token budgeted optimization problem balancing relevance and redundancy \citep{peng2025adagres}. These approaches reason about capacity and redundancy, but they do not explicitly model multi-hop information gaps or use targeted micro-queries to repair missing evidence.

AdaGATE combines these two perspectives. Like SEAL-RAG, it is a training-free controller that performs explicit gap-aware evidence repair with an entity centric ledger. Like Adaptive-$k$ and AdaGReS, it reasons about context efficiency under limited budgets. Its key difference is to integrate gap-aware repair, question-aware fallback retrieval, redundancy-aware utility scoring, and adaptive capacity control within a single evidence selection procedure for multi-hop RAG.

\section{Method}
\label{sec:method}

We formulate multi-hop RAG under deployment constraints as a token constrained evidence repair problem. Given a query $q$, a corpus $\mathcal{D}$, and a global token budget $B$, the goal is to assemble a compact evidence set that supports multi-hop reasoning while avoiding redundant or misleading passages. AdaGATE is a training-free controller built on top of a fixed retriever and generator. At each iteration $t$, it maintains an evidence set $E_t$, an entity centric ledger $U_t$, and a set of unresolved information gaps $G_t$. Relative to SEAL-RAG \citep{lahmy2025sealrag}, AdaGATE makes three changes: it replaces fixed-$k$ evidence selection with constrained token selection, adds a question-aware fallback channel to gap targeted retrieval, and uses adaptive utility capacity control to avoid overfilling the context with low value passages.

\subsection{Gap-Aware Retrieval and Evidence State}

Let $\mathcal{C}_t$ denote the candidate pool retrieved at iteration $t$. Each passage $c \in \mathcal{C}_t$ has token length $\ell(c)$, and the final evidence set must satisfy
\begin{equation}
\sum_{c \in E_t} \ell(c) \le B.
\label{eq:budget}
\end{equation}

Following SEAL-RAG, AdaGATE uses two LLM-based primitives: (1) ledger extraction, which summarizes the current evidence set into structured entity--relation--value tuples with confidence scores, and (2) gap specification, which identifies missing facts needed to answer the question \citep{lahmy2025sealrag}. We treat these as black-box components and focus on how AdaGATE uses them to guide retrieval and evidence selection under a limited context budget.

For each gap $g \in G_t$, AdaGATE generates one or more targeted micro-queries. To improve robustness when gap extraction is noisy or overly abstract, it also generates a small set of question-anchored fallback queries derived directly from $q$. The union of gap-aware and question-aware queries is sent to the retriever to form the next candidate pool $\mathcal{C}_t$. This design allows the controller to continue exploring useful evidence even when the current gap representation is incomplete.

\subsection{Utility-Based Evidence Scoring}

Given the current query, ledger, gaps, and evidence state, AdaGATE assigns each candidate passage $c \in \mathcal{C}_t$ a scalar utility score
\begin{align}
S_t(c) &= \lambda_1 \,\mathrm{GapCov}(c,G_t)
          + \lambda_2 \,\mathrm{Corr}(c,U_t) \nonumber \\
       &\quad + \lambda_3 \,\mathrm{Nov}(c,U_t)
          - \lambda_4 \,\mathrm{Red}(c,E_t) \nonumber \\
       &\quad + \lambda_5 \,\mathrm{Rel}_Q(c,q),
\label{eq:utility}
\end{align}

The five terms capture complementary roles in multi-hop evidence assembly. $\mathrm{GapCov}(c,G_t)$ rewards passages that directly address unresolved gaps. $\mathrm{Corr}(c,U_t)$ rewards support for low-confidence facts already present in the ledger. $\mathrm{Nov}(c,U_t)$ favors passages that contribute new entities or relations rather than repeating lateral information. $\mathrm{Red}(c,E_t)$ penalizes candidates that are highly similar to evidence already selected. Finally, $\mathrm{Rel}_Q(c,q)$ measures direct relevance to the original question and acts as a fallback signal when gap extraction is noisy. Compared with SEAL-RAG, the most important additions are the explicit redundancy penalty and the question-aware relevance term, which together make the controller more robust under noisy or redundant retrieval.

\subsection{Token-Constrained Selection with Adaptive Capacity}

AdaGATE does not fix the number of passages passed to the generator. Instead, it selects evidence under the token budget in Eq.~\ref{eq:budget}, allowing the final evidence set size to vary with passage length and utility. In practice, AdaGATE uses the utility score in Eq.~\ref{eq:utility} as a surrogate for marginal value and greedily assembles a compact evidence set from the highest scoring candidates.

To avoid filling the available budget with many mediocre passages, AdaGATE estimates an effective capacity from the utility distribution. Let
\[
S_t^{(1)} \ge S_t^{(2)} \ge \dots \ge S_t^{(M)}
\]
denote candidate utilities sorted in descending order, and define adjacent drops
\[
\Delta_i = S_t^{(i)} - S_t^{(i+1)}.
\]
AdaGATE chooses the largest drop
\[
i^\star = \arg\max_i \Delta_i
\]
and sets
\[
K_t^{\text{eff}} = i^\star + B_{\text{buf}},
\]
where $B_{\text{buf}}=2$ is a small buffer. The largest utility drop separates a high-value prefix from a lower-value tail; AdaGATE prioritizes candidates within the top $K_t^{\text{eff}}$ range and greedily selects from them while enforcing the global token budget.

AdaGATE iterates over four stages: \textbf{\textit{extract}}, \textbf{\textit{search}}, \textbf{\textit{score}}, and \textbf{\textit{replace}}. It first extracts the current ledger and unresolved gaps from $E_t$, then retrieves new candidates using both gap-aware and question-aware queries, scores candidates with Eq.~\ref{eq:utility} and estimates the effective capacity, and finally updates the evidence set by replacing lower utility passages with higher utility candidates while respecting the token budget. The process stops when no useful repair remains, no meaningful gaps are identified, or a maximum number of repair iterations is reached. After termination, the final evidence set is concatenated with the question and passed to the generator.

\section{Experimental Setup}
\label{sec:experimental-setup}

\subsection{Dataset and Retrieval Setup}
\label{subsec:datasets}

We evaluate on HotpotQA \citep{yang2018hotpotqa}, a multi-hop QA benchmark over Wikipedia in which each question is associated with two supporting paragraphs and additional distractor passages. We use the distractor setting, which provides both relevant and irrelevant evidence and is therefore suitable for studying evidence selection under imperfect retrieval.

All controllers share the same retrieval infrastructure. We use a single Pinecone index built from the first 1{,}000 HotpotQA validation examples, yielding 10{,}919 document chunks embedded with OpenAI \texttt{text-embedding-3-small}. Evaluation is conducted on $N=200$ validation examples. Because all methods retrieve from the same index and use the same embedding model, observed differences reflect evidence control rather than changes in the retriever.

\subsection{Stress-Test Retrieval Conditions}
\label{subsec:stress-tests}

To evaluate robustness beyond the clean benchmark setting, we construct controlled perturbations of the candidate pools while keeping the question, gold answer, and supporting facts unchanged. Each condition is indexed into a separate Pinecone namespace.

\paragraph{Noise injection.}
We construct a mixed noise pool by combining two perturbation types applied to the original passages: syntax distortion (word order scrambling, spelling corruption, and partial truncation) and cross-query injection (irrelevant passages sampled from other examples). The noise ratio is fixed at $\rho_{\text{noise}} = 0.5$, increasing the candidate pool from 10 to 20 passages per example.

\paragraph{Redundancy injection.}
We augment each example with paraphrastic or partially overlapping variants of the gold supporting passages. The redundancy ratio is fixed at $\rho_{\text{red}} = 0.5$, increasing the total pool to approximately 19{,}854 indexed documents. Unlike the noise condition, these passages are topically relevant but largely non-complementary, testing whether a controller can avoid spending budget on repeated evidence.

\subsection{Models and Baselines}
\label{subsec:models}

All methods use the same retriever, generator, and evaluation setup. The retriever is OpenAI \texttt{text-embedding-3-small} with Pinecone, retrieving $k=3$ passages per query. The generator is \texttt{gpt-4o-mini}, treated as a black-box backbone without finetuning.

We compare five controller settings:
\begin{itemize}
  \item \textbf{Basic RAG}: retrieve the top-$k$ passages and generate directly.
  \item \textbf{Self-RAG} \citep{asai2023selfrag}: retrieval aware self-reflective generation.
  \item \textbf{Adaptive-$k$} \citep{taguchi2025adaptivek}: query specific passage count based on similarity score gaps.
  \item \textbf{SEAL-RAG} \citep{lahmy2025sealrag}: entity centric gap-repair controller, evaluated with $L \in \{1,3\}$ repair iterations.
  \item \textbf{\textsc{AdaGATE}}: our gap-aware, token-constrained controller, evaluated with $L \in \{1,3\}$ and global token budget $B=3000$.
\end{itemize}

This comparison spans fixed size, adaptive size, and repair based evidence controllers under a shared retrieval and generation pipeline.

\subsection{Evaluation Metrics}
\label{subsec:evaluation}

We evaluate along four dimensions: answer correctness, evidence quality, grounding quality, and token efficiency.

\paragraph{Answer correctness.}
We use \texttt{gpt-4o} as a judge to compare predicted answers against gold answers semantically and assign a binary correctness label. Using a stronger judge than the generator reduces self-evaluation bias \citep{saadfalcon2023ares}.

\paragraph{Evidence quality.}
We report precision, recall, and F1 against the gold supporting document titles annotated in HotpotQA. Retrieved titles are extracted from chunk headings and deduplicated before comparison.

\paragraph{Grounding quality.}
We follow ARES \citep{saadfalcon2023ares} and report Context Relevance (CR), Answer Faithfulness (AF), and Answer Relevance (AR), again using \texttt{gpt-4o} as the judge. These metrics complement title based retrieval measures with generation-aware judgments over the retrieved evidence and final answer.

\paragraph{Token efficiency.}
We measure average input tokens per query, average number of documents passed to the generator, and tokens per correctly answered question, enabling direct cost--quality comparisons across controllers.

All judges are blinded to controller identity and see only the question, answer, and retrieved passages. We use fixed prompts throughout.

\section{Results}

We report results across seven controllers and three conditions. Full numerical results are in Table~\ref{tab:full_results} in the Appendix.

\subsection{Evidence Quality and Answer Correctness on Clean Data}
\label{subsec:accuracy}

\begin{figure*}[!t]
    \centering
    \includegraphics[width=0.95\textwidth]{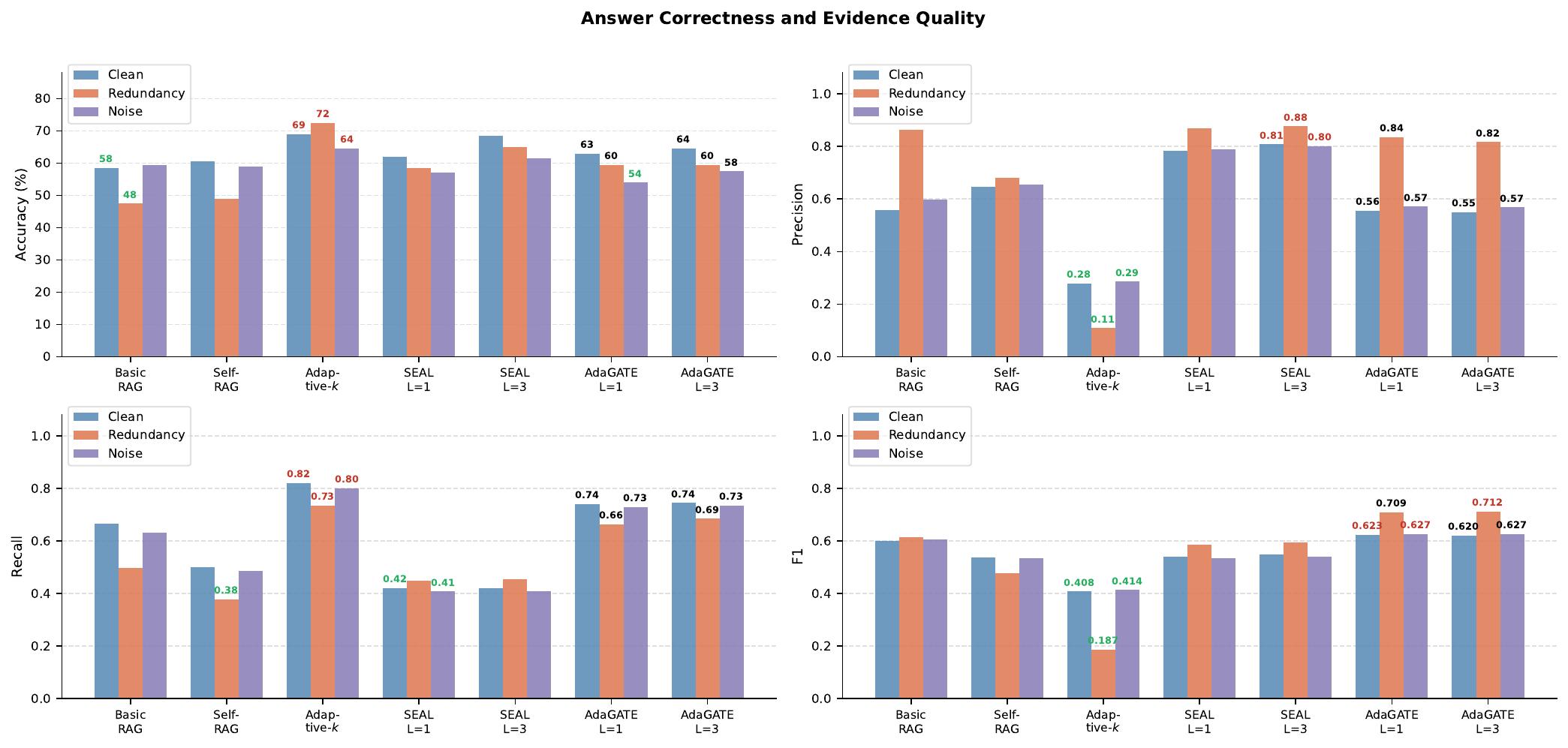}
    \caption{Answer correctness and evidence quality. Red = best per condition; green = worst; black = AdaGATE otherwise. AdaGATE achieves the highest F1 across all three conditions.}
    \label{fig:accuracy_f1}
\end{figure*}

Figure~\ref{fig:accuracy_f1} reveals a consistent tension between precision and recall across all controllers in clean data.

\paragraph{Add-only controllers.}
Adaptive-$k$ achieves the highest accuracy (69.0\%) but at the cost of severely collapsed precision (P=0.278). Its high recall (R=0.820) arises because it retrieves an average of 8.6 documents per query, admitting many irrelevant passages alongside the gold supporting ones. Basic RAG is the most balanced passive baseline (Acc=58.5\%, F1=0.601). Self-RAG's document grading mechanism occasionally rejects all retrieved passages, leaving the generator with an empty evidence set and explaining its lower recall relative to Basic RAG.

\paragraph{SEAL-RAG: implementation versus paper.}
SEAL-RAG achieves the highest precision (P=0.808 at $L{=}3$), but recall remains low and flat at approximately 0.42 regardless of the condition or repair iterations. This behavior is directly traceable to a discrepancy between the SEAL-RAG paper and its implementation. The paper describes a utility-based ranking mechanism for evidence selection; the actual implementation replaces this with an LLM entity selection step that instructs the model to ``choose the fewest entities needed to answer.'' In practice, this causes the model to select a single entity on the majority of questions, resulting in an average of only 1.0 documents passed to the generator. High precision follows naturally from single document selection, but this context systematically misses bridge facts on multi-hop questions requiring evidence from two or more sources. Increasing repair iterations from $L{=}1$ to $L{=}3$ improves accuracy modestly (62.0\%$\to$68.5\%) but does not resolve the fundamental recall ceiling.

\paragraph{AdaGATE.}
AdaGATE addresses this recall bottleneck by passing an average of 2.8--3.0 documents to the generator, assembling a richer evidence set through its adaptive capacity mechanism while suppressing redundant passages via the utility scoring function. The result is the highest F1 on clean data (62.3\% at $L{=}1$), outperforming SEAL-RAG ($L{=}1$) by 8.2 F1 points.

\subsection{Token Efficiency}
\label{subsec:token}

\begin{figure*}[!t]
    \centering
    \includegraphics[width=0.95\textwidth]{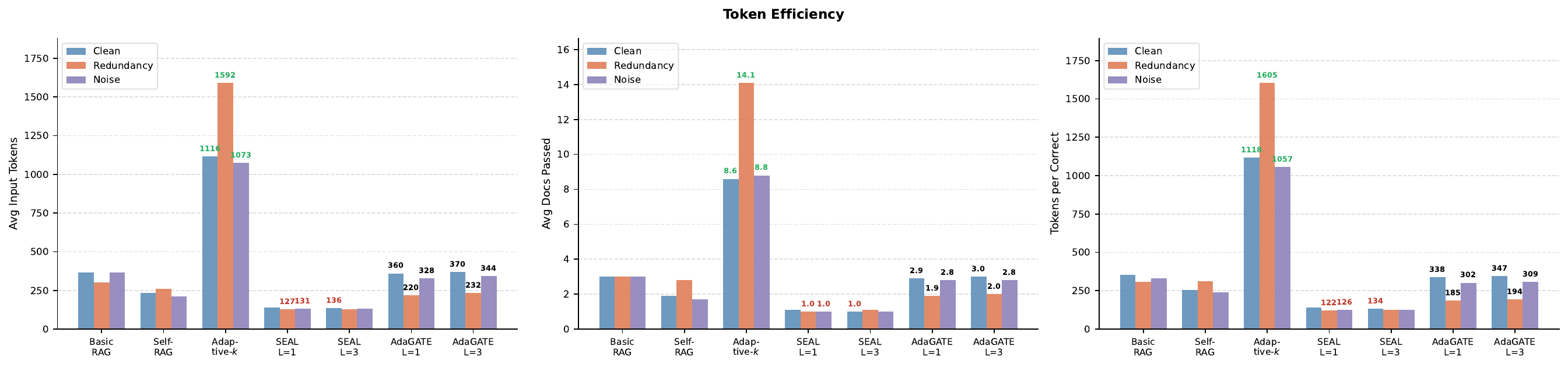}
    \caption{Token efficiency across controllers and conditions. Red = most efficient; green = least efficient. Adaptive-$k$ is consistently the least token-efficient controller.}
    \label{fig:token_efficiency}
\end{figure*}

Figure~\ref{fig:token_efficiency} shows that controllers differ by more than an order of magnitude in token usage. Adaptive-$k$ is the least efficient in every condition, consuming 1,116 tokens on clean data and 1,592 under redundancy, with tokens-per-correct of 1,118---roughly 3.3$\times$ the cost of AdaGATE (338) and 8.3$\times$ SEAL-RAG (134). SEAL-RAG is the most token efficient at 136--140 tokens, a direct consequence of single document selection, but at the cost of recall established in Section~\ref{subsec:accuracy}.

AdaGATE uses 360 tokens on clean data, 2.6$\times$ fewer than Adaptive-$k$ while achieving substantially higher F1. Under redundancy, token usage drops adaptively to 220--232 tokens as its redundancy penalty concentrates the evidence set on fewer, higher utility passages---simultaneously reducing cost and improving evidence quality.

\subsection{ARES Grounding Scores}
\label{subsec:ares}

\begin{figure*}[!t]
    \centering
    \includegraphics[width=0.95\textwidth]{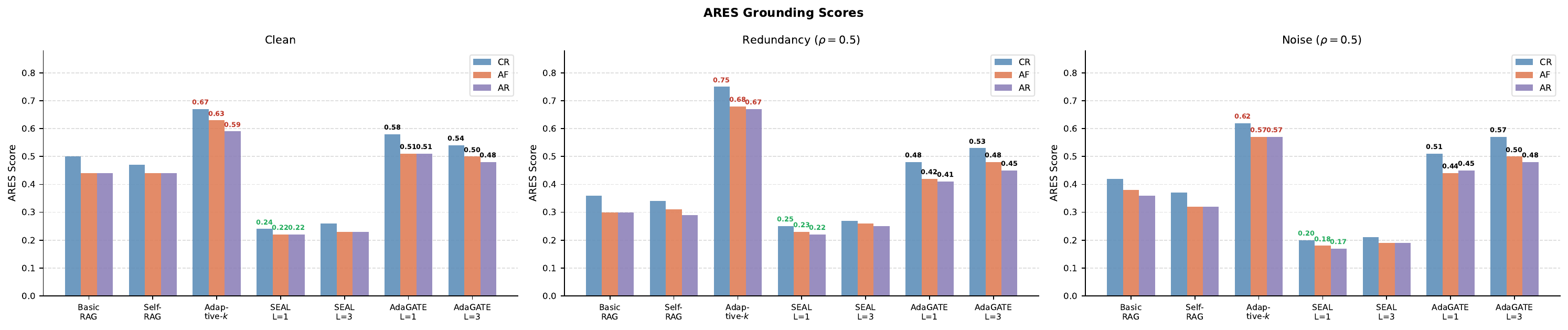}
    \caption{ARES grounding scores (CR = Context Relevance, AF = Answer Faithfulness, AR = Answer Relevance). Red = best; green = worst. SEAL-RAG scores are consistently the lowest despite high retrieval precision.}
    \label{fig:ares}
\end{figure*}

Figure~\ref{fig:ares} reveals a striking divergence between retrieval precision and ARES grounding scores. Adaptive-$k$ leads on all three ARES dimensions (CR=0.67, AF=0.63, AR=0.59 on clean) despite collapsed retrieval precision, because its large context of 8--14 documents inflates faithfulness scores by increasing the probability that at least one passage supports the generated answer. This highlights a fundamental limitation of document conditioned evaluation under retrieval with high recall: ARES scores reflect context availability rather than evidence precision.

SEAL-RAG produces the lowest ARES scores across all conditions (CR=0.24--0.26, AF=0.22--0.23 on clean). When the single selected document fails to contain the complete reasoning chain, the generator abstains rather than hallucinating, and the ARES judge penalizes abstentions as not faithful---explaining why SEAL-RAG's CR substantially exceeds its AF.

AdaGATE achieves intermediate ARES scores (CR=0.58, AF=0.51, AR=0.51 on clean) substantially higher than SEAL-RAG. The gap between AdaGATE and SEAL-RAG on ARES is larger than the gap on retrieval F1, suggesting that multi-document assembly provides meaningfully better grounding support and reduces abstention rates.

\subsection{Robustness Under Stress-Test Conditions}
\label{subsec:stress}

\paragraph{Redundancy injection.}
Under redundancy ($\rho{=}0.5$), accuracy drops sharply for Basic RAG (58.5\%$\to$47.5\%) and Self-RAG (60.5\%$\to$49.0\%). Adaptive-$k$'s accuracy counterintuitively increases to 72.5\% as its high-recall strategy benefits from a pool containing many topically relevant paraphrases, but precision collapses to P=0.109 and ARES scores drop sharply. SEAL-RAG accuracy is relatively stable (58.5\%--65.0\%) with high precision (P=0.868--0.876), but recall drops further (R=0.448--0.455) as the retriever must surface gold documents from a pool twice the size, and ARES scores stagnate.

AdaGATE's F1 improves substantially under redundancy (62.3\%$\to$71.2\% at $L{=}3$, and 62.3\%$\to$70.9\% at $L{=}1$), driven by the redundancy penalty term. When the candidate pool is dominated by paraphrase variants, the novelty and redundancy terms jointly suppress their utility scores, causing AdaGATE to select fewer but more complementary documents. Token usage drops adaptively from 360 to 220 tokens, and ARES scores remain stable at $L{=}3$ (CR=0.53, AF=0.48), while Basic RAG's drop sharply (CR=0.50$\to$0.36).

\paragraph{Noise injection.}
Under noise, all controllers degrade in accuracy. Noise passages retain high embedding similarity because syntax distortion does not fundamentally alter semantic representations, causing them to appear within the top-$k$. AdaGATE's accuracy drops to 54.0\% ($L{=}1$), its weakest result, but F1 (62.7\%) matches clean performance, indicating that the utility scoring function partially penalizes corrupted passages through lower GapCov and RelQ scores. Under noise at $L{=}3$, AdaGATE achieves the highest CR among non-Adaptive-$k$ controllers (0.57), suggesting that utility based filtering partially maintains context relevance even under retrieval degradation.

\subsection{Pipeline-Level Analysis}
\label{subsec:pipeline}

Table~\ref{tab:case_study} presents three representative examples from clean evaluation logs, spanning three distinct behavioral patterns: systematic single document recall loss, micro-query failure without fallback, and over conservative gap detection.

\begin{table*}[!t]
\centering
\small
\setlength{\tabcolsep}{6pt}
\renewcommand{\arraystretch}{1.8}
\begin{tabular}{p{2.6cm} p{5.8cm} p{5.8cm}}
\toprule
\textbf{Question} & \textbf{SEAL-RAG ($L{=}1$)} & \textbf{AdaGATE ($L{=}1$)} \\
\midrule
\textit{Q1: Were Scott Derrickson and Ed Wood of the same nationality?}
&
\textbf{Retrieve:} Dispatches 3 docs, extracts 3 entities. Micro query: None. \newline
\textbf{Repair:} yes --- both nationalities confirmed in retrieved docs. \newline
\textbf{Rank:} entity selection collapses to 1 doc; Ed Wood passage discarded. \newline
\textbf{Generate:} docs=1, tokens=97, P=1.00, R=0.50, F1=0.67. \newline
\cmark\ ``Yes, both American.''
&
\textbf{Retrieve:} Dispatches 2 docs, extracts 2 entities. \newline
\textbf{Repair:} yes. Utility scores [0.19, 0.15]. \newline
\textbf{Rank:} $K_{\text{eff}}=2$, both gold docs passed to generator. \newline
\textbf{Generate:} docs=2, tokens=136, P=1.00, R=1.00, F1=1.00. \newline
\cmark\ ``Yes, both American.'' \\
\midrule
\textit{Q3: What science fantasy YA series has companion books about enslaved worlds?}
&
\textbf{Retrieve:} Dispatches 3 docs, extracts 7 entities. Repair: no --- Animorphs companion books not confirmed. \newline
\textbf{Micro query:} ``young adult science fantasy series companion books enslaved worlds alien species.'' \texttt{WARNING: No documents to extract entities from!} No new docs retrieved. \newline
\textbf{Loop limit reached.} Repair: no (unchanged). \newline
\textbf{Generate:} docs=1, tokens=128, P=1.00, R=0.50, F1=0.67. \newline
\xmark\ ``I don't know.'' \quad gt: ``Animorphs''
&
\textbf{Retrieve:} Dispatches 2 docs, extracts 4 entities. Utility scores [0.23, 0.19]. Repair: no --- companion books not confirmed. \newline
\textbf{Micro query:} $H_{\text{gap}}$ (3 docs) $\cup$ $H_q$ fallback (3 docs). 1 new doc dispatched for entity extraction; new entity about Hork-Bajir Chronicles retrieved. Total: 9 entities, 3 docs. Utility scores [0.32, 0.26, 0.23]. \newline
\textbf{Repair:} yes --- Animorphs companion books confirmed. $K_{\text{eff}}=3$, 3 docs passed. \newline
\textbf{Generate:} docs=3, tokens=500, P=0.67, R=1.00, F1=0.80. \newline
\cmark\ ``Animorphs series'' \\
\midrule
\textit{Q21: Which other Mexican F1 driver held the podium besides Sergio Pérez?}
&
\textbf{Retrieve:} Dispatches 3 docs, extracts 4 entities including Pedro Rodríguez. Repair: no --- podium finish not explicitly confirmed. \newline
\textbf{Micro query:} ``Mexican Formula One drivers podium finishes history.'' Returns 1 new doc; 0 new entities extracted (entity count unchanged at 4). \newline
\textbf{Repair:} no --- evidence still insufficient. \newline
\textbf{Loop limit reached.} Generator infers from partial evidence. \newline
\textbf{Generate:} docs=1, tokens=100, P=1.00, R=0.50, F1=0.67. \newline
\cmark\ ``Pedro Rodríguez'' \quad gt: ``Pedro Rodríguez''
&
\textbf{Retrieve:} Dispatches 2 docs, extracts 4 entities. Utility scores [0.24, 0.21]. Repair: no --- podium finish not confirmed. \newline
\textbf{Micro query:} $H_{\text{gap}} \cup H_q$ retrieves 6 docs; 1 dispatched for entity extraction, 0 new entities (count unchanged at 4). Utility scores [0.38, 0.26, 0.25]. \newline
\textbf{Repair:} no --- gap still open across both iterations. \newline
\textbf{Loop limit reached.} Generator abstains. \newline
\textbf{Generate:} docs=3, tokens=511, P=0.67, R=1.00, F1=0.80. \newline
\xmark\ ``I don't know.'' \quad gt: ``Pedro Rodríguez'' \\
\bottomrule
\end{tabular}
\caption{Pipeline-level comparison of SEAL-RAG and AdaGATE on three representative HotpotQA examples (clean, $L{=}1$). Q1 shows SEAL-RAG's systematic single-document selection and its recall cost even when both controllers answer correctly. Q3 shows AdaGATE's $H_q$ fallback channel recovering from a micro-query that returns zero new documents for SEAL-RAG. Q21 shows AdaGATE's conservative gap detection causing abstention while SEAL-RAG's less stringent sufficiency check produces the correct answer from partial evidence.}
\label{tab:case_study}
\end{table*}

Q1 illustrates the most common behavioral pattern: SEAL-RAG's entity selection discards the Ed Wood passage despite answering correctly, while AdaGATE passes both gold documents. This recall gap is inconsequential for simple comparisons but compounds on questions requiring multi-hop bridge reasoning. Q3 shows AdaGATE's $H_q$ fallback recovering from a failed micro-query that returns zero new documents for SEAL-RAG, a failure mode that occurs whenever the gap specification is too abstract to match any indexed passage. Q21 exposes the opposite failure: AdaGATE's stringent gap detection causes abstention when indirect evidence would suffice, while SEAL-RAG's less conservative sufficiency check generates the correct answer from partial evidence---illustrating a fundamental tradeoff between hallucination prevention and abstention rate.

\paragraph{Summary.}
AdaGATE achieves \textbf{better precision-recall balance} than fixed-$k$ and add-only baselines \textbf{at comparable token cost}, and \textbf{more robust} under stress-test conditions. The pipeline analysis further reveals that SEAL-RAG's apparent precision advantage stems from an implementation that departs from its stated design, while AdaGATE's multi-document assembly and fallback retrieval provide more robust evidence coverage at the cost of occasional over conservative abstention.

\section{Discussion}

\paragraph{Hypothesis confirmation.}
Both central hypotheses were confirmed. AdaGATE achieves the highest F1 across all three conditions (62.3\% clean, 71.2\% redundancy, 62.7\% noise) while using 2.6$\times$ fewer tokens than Adaptive-$k$. F1 under redundancy improves relative to clean, and F1 under noise matches clean performance despite accuracy dropping, confirming that the utility scoring function maintains evidence completeness even when individual passage quality deteriorates.

\paragraph{Relation to prior work.}
The precision-recall tradeoff between SEAL-RAG and Adaptive-$k$ mirrors the broader tension between context precision and answer coverage in the RAG literature \citep{cuconasu2024powerofnoise}. We additionally identify a concrete implementation gap in SEAL-RAG: its stated utility based ranking is replaced in practice by an LLM entity selection step that collapses the evidence set to one document by construction. The ARES analysis further reveals that Adaptive-$k$'s inflated scores reflect context availability rather than evidence quality, consistent with concerns about document conditioned evaluation \citep{saadfalcon2023ares}.

\paragraph{Limitations and future work.}
The current evaluation uses $k=3$ retrieval, which limits the candidate pool and prevents the token budget from binding; larger $k$ or longer repair chains would engage the budget more actively. Evaluation is also restricted to HotpotQA's distractor setting, and performance on web-scale retrieval may differ. The utility weights $(\lambda_1,\ldots,\lambda_5)$ were set heuristically and could be learned from supervision. Finally, AdaGATE's conservative gap detection leads to over abstention on questions where indirect evidence would suffice; calibrating the sufficiency threshold to question type is a natural direction for future work.

\section{Conclusion}

We presented AdaGATE, a training-free evidence controller for multi-hop RAG that combines entity centric gap tracking, a five dimensional utility scoring function, adaptive capacity control, and a question anchored fallback retrieval channel. Evaluated on HotpotQA across clean, redundancy and noise injected conditions, AdaGATE achieves the highest evidence F1 in all three settings while using 2.6$\times$ fewer tokens than Adaptive-$k$. A pipeline-level analysis reveals that SEAL-RAG's precision advantage is an artifact of an implementation that departs from its stated design, and that AdaGATE's multi-document assembly provides more robust evidence coverage at the cost of occasional over conservative abstention. These results support viewing multi-hop RAG as a token-budgeted evidence repair problem and suggest that explicit gap modeling, redundancy-aware scoring, and fallback retrieval together constitute a principled approach to robust evidence assembly under realistic deployment constraints.

\section*{Acknowledgments}
This work was carried out as a course project for DS-GA / LING-GA 1012 (Natural Language Understanding and Computational Semantics) at New York University, Spring 2026. We thank the course instructor, Prof.~Tal Linzen, and the teaching staff for their feedback during in-semester presentations. We also thank the authors of SEAL-RAG \citep{lahmy2025sealrag} for releasing their code, which served as the starting point for our implementation.

\bibliography{custom}

\appendix

\begin{appendices}
\section{Full Evaluation Results}
\label{sec:appendix}

Table~\ref{tab:full_results} reports complete numerical results for all controllers across all conditions and metrics.

\begin{table*}[!t]
\centering
\small
\setlength{\tabcolsep}{3pt}
\renewcommand{\arraystretch}{1.6}
\resizebox{\textwidth}{!}{
\begin{tabular}{l l cccc cccc cccc}
\toprule
& & \multicolumn{4}{c}{\textbf{Clean}}
  & \multicolumn{4}{c}{\textbf{Redundancy ($\rho{=}0.5$)}}
  & \multicolumn{4}{c}{\textbf{Noise ($\rho{=}0.5$)}} \\
\cmidrule(lr){3-6}\cmidrule(lr){7-10}\cmidrule(lr){11-14}
& & Acc & P & R & F1 & Acc & P & R & F1 & Acc & P & R & F1 \\
\midrule
\multirow{7}{*}{\rotatebox{90}{\textbf{Controller}}}
& Basic RAG
  & 58.5 & .559 & .665 & .601
  & 47.5 & .862 & .498 & .615
  & 59.5 & .598 & .630 & .605 \\
& Self-RAG
  & 60.5 & .645 & .500 & .536
  & 49.0 & .681 & .378 & .476
  & 59.0 & .655 & .485 & .534 \\
& Adaptive-$k$
  & \textbf{69.0} & .278 & \textbf{.820} & .408
  & \textbf{72.5} & .109 & \textbf{.735} & .187
  & \textbf{64.5} & .286 & \textbf{.800} & .414 \\
& SEAL-RAG ($L{=}1$)
  & 62.0 & .784 & .420 & .541
  & 58.5 & .868 & .448 & .587
  & 57.0 & .790 & .408 & .535 \\
& SEAL-RAG ($L{=}3$)
  & 68.5 & \textbf{.808} & .420 & .549
  & 65.0 & \textbf{.876} & .455 & .595
  & 61.5 & \textbf{.800} & .410 & .540 \\
& \textsc{AdaGATE} ($L{=}1$, ours)
  & 63.0 & .555 & .740 & \textbf{.623}
  & 59.5 & .836 & .663 & .709
  & 54.0 & .571 & .728 & \textbf{.627} \\
& \textsc{AdaGATE} ($L{=}3$, ours)
  & 64.5 & .549 & .745 & .620
  & 59.5 & .817 & .685 & \textbf{.712}
  & 57.5 & .568 & .735 & \textbf{.627} \\
\midrule
\multicolumn{14}{l}{\textit{Token efficiency (avg\_tokens / avg\_docs / tokens\_per\_correct)}} \\[4pt]
& Basic RAG
  & \multicolumn{4}{c}{364 / 3.0 / 352}
  & \multicolumn{4}{c}{300 / 3.0 / 306}
  & \multicolumn{4}{c}{364 / 3.0 / 330} \\
& Self-RAG
  & \multicolumn{4}{c}{234 / 1.9 / 254}
  & \multicolumn{4}{c}{259 / 2.8 / 311}
  & \multicolumn{4}{c}{211 / 1.7 / 240} \\
& Adaptive-$k$
  & \multicolumn{4}{c}{1116 / 8.6 / 1118}
  & \multicolumn{4}{c}{1592 / 14.1 / 1605}
  & \multicolumn{4}{c}{1073 / 8.8 / 1057} \\
& SEAL-RAG ($L{=}1$)
  & \multicolumn{4}{c}{140 / 1.1 / 139}
  & \multicolumn{4}{c}{\textbf{127} / \textbf{1.0} / \textbf{122}}
  & \multicolumn{4}{c}{\textbf{131} / \textbf{1.0} / \textbf{126}} \\
& SEAL-RAG ($L{=}3$)
  & \multicolumn{4}{c}{\textbf{136} / \textbf{1.0} / \textbf{134}}
  & \multicolumn{4}{c}{128 / 1.1 / 125}
  & \multicolumn{4}{c}{\textbf{131} / \textbf{1.0} / 127} \\
& \textsc{AdaGATE} ($L{=}1$, ours)
  & \multicolumn{4}{c}{360 / 2.9 / 338}
  & \multicolumn{4}{c}{220 / 1.9 / 185}
  & \multicolumn{4}{c}{328 / 2.8 / 302} \\
& \textsc{AdaGATE} ($L{=}3$, ours)
  & \multicolumn{4}{c}{370 / 3.0 / 347}
  & \multicolumn{4}{c}{232 / 2.0 / 194}
  & \multicolumn{4}{c}{344 / 2.8 / 309} \\
\midrule
\multicolumn{14}{l}{\textit{ARES scores (CR / AF / AR)}} \\[4pt]
& Basic RAG
  & \multicolumn{4}{c}{.50 / .44 / .44}
  & \multicolumn{4}{c}{.36 / .30 / .30}
  & \multicolumn{4}{c}{.42 / .38 / .36} \\
& Self-RAG
  & \multicolumn{4}{c}{.47 / .44 / .44}
  & \multicolumn{4}{c}{.34 / .31 / .29}
  & \multicolumn{4}{c}{.37 / .32 / .32} \\
& Adaptive-$k$
  & \multicolumn{4}{c}{\textbf{.67} / \textbf{.63} / \textbf{.59}}
  & \multicolumn{4}{c}{\textbf{.75} / \textbf{.68} / \textbf{.67}}
  & \multicolumn{4}{c}{\textbf{.62} / \textbf{.57} / \textbf{.57}} \\
& SEAL-RAG ($L{=}1$)
  & \multicolumn{4}{c}{.24 / .22 / .22}
  & \multicolumn{4}{c}{.25 / .23 / .22}
  & \multicolumn{4}{c}{.20 / .18 / .17} \\
& SEAL-RAG ($L{=}3$)
  & \multicolumn{4}{c}{.26 / .23 / .23}
  & \multicolumn{4}{c}{.27 / .26 / .25}
  & \multicolumn{4}{c}{.21 / .19 / .19} \\
& \textsc{AdaGATE} ($L{=}1$, ours)
  & \multicolumn{4}{c}{.58 / .51 / .51}
  & \multicolumn{4}{c}{.48 / .42 / .41}
  & \multicolumn{4}{c}{.51 / .44 / .45} \\
& \textsc{AdaGATE} ($L{=}3$, ours)
  & \multicolumn{4}{c}{.54 / .50 / .48}
  & \multicolumn{4}{c}{.53 / .48 / .45}
  & \multicolumn{4}{c}{.57 / .50 / .48} \\
\bottomrule
\end{tabular}
}
\caption{Full evaluation results (HotpotQA distractor, $k=3$, $N=200$).
         Accuracy (\%) judged by GPT-4o. P = Precision, R = Recall, F1 against gold titles.
         Token efficiency: avg\_tokens / avg\_docs / tokens\_per\_correct.
         CR = Context Relevance, AF = Answer Faithfulness, AR = Answer Relevance
         (ARES UES/IDP, GPT-4o). \textbf{Bold} = best per metric per condition.}
\label{tab:full_results}
\end{table*}
\end{appendices}

\end{document}